\RequirePackage{amsmath}
\documentclass[runningheads]{llncs}

\usepackage{multirow}
\usepackage{graphicx}
\usepackage{hyperref}
\usepackage{xcolor}
\usepackage{tabularx}
\usepackage{mathtools}
\usepackage{booktabs}

\usepackage[T1]{fontenc}
\usepackage{graphicx}
\usepackage{color}

\begin{document}
\title{Context-Aware Content Moderation for German Newspaper Comments}
%
%
\author{Felix Krejca\orcidID{0009-0002-0549-0865}\inst{1,2} \and Tobias Kietreiber\orcidID{0009-0003-4396-1135}\inst{1}\and Alexander Buchelt\orcidID{0000-0003-3851-6320}\inst{1} \and Sebastian Neumaier\orcidID{0000-0002-9804-4882}\inst{1}}

\authorrunning{F. Krejca, et al.}
%
\institute{St.\ P\"olten University of Applied Sciences, Austria\\
\email{firstname.lastname@fhstp.ac.at} \and Der Standard, Vienna, Austria}
\maketitle              
\begin{abstract}
The increasing volume of online discussions requires advanced automatic content moderation to maintain responsible discourse. While hate speech detection on social media is well-studied, research on German-language newspaper forums remains limited. Existing studies often neglect platform-specific context, such as user history and article themes. This paper addresses this gap by developing and evaluating binary classification models for automatic content moderation in German newspaper forums, incorporating contextual information. Using LSTM, CNN, and ChatGPT-3.5 Turbo, and leveraging the One Million Posts Corpus from the Austrian newspaper Der Standard, we assess the impact of context-aware models. Results show that CNN and LSTM models benefit from contextual information and perform competitively with state-of-the-art approaches. In contrast, ChatGPT's zero-shot classification does not improve with added context and underperforms. 

\keywords{Automatic Content Moderation \and Context-Aware Text Classification \and Hate Speech Detection.}
\end{abstract}
\section{Introduction}

With the rising quantity of text data in different kinds of media, the theoretical understanding and the practical implications of automatic content moderation based on machine learning techniques is more important than ever. 
\setcounter{footnote}{0}
Recent developments, such as the EU Digital Services Act (DSA) and other regulatory efforts, highlight the growing pressure on platforms to implement transparent and effective content moderation systems.\footnote{\url{https://digital-strategy.ec.europa.eu/en/policies/dsa-impact-platforms\#ecl-inpage-lo5phy1n}, last accessed 2025-02-05} 
Manual moderation, while thorough, often exposes moderators to distressing material, leading to significant emotional and psychological challenges. Studies have documented that content moderators frequently encounter graphic and harmful content, resulting in conditions such as post-traumatic stress disorder, anxiety, and depression \cite{spence2023psychological}. AI-driven moderation has the potential to diminish the emotional toll on moderators by handling straightforward decisions, thereby reducing their exposure to harmful material. 

While much of the recent research has focused on detecting hate speech and offensive language in social media (e.g., \cite{alkomah2022,malik2024,mullah2021}), limited studies have explored automated content moderation within German-language newspaper forums. Moreover, most existing research in this area does not utilize platform-specific context information, such as user posting history or details about the article under which a comment is posted (e.g., \cite{AssenmacherNMSR21,schabus2018,schabus2017,yadav2021,pachinger2024,krenn2024,keller2025}). 

This paper contributes to this branch of research by providing novel approaches for binary automatic content moderation with new ways of using platform-specific contextual information, to decide if a comment should stay online (0) or be removed (1), for German newspaper comments, using LSTM, CNN and LLM (ChatGPT: GPT-3.5-Turbo)\footnote{\url{https://platform.openai.com/docs/models/gpt-3-5-turbo/}, last accessed 2025-02-06} models. We compare the models and prompts to the state-of-the-art classification models for automatic content moderation and offensive language classification of German newspaper comments (e.g., \cite{AssenmacherNMSR21,schabus2017,vidgen2021}).
We address the following central research question:
\textit{How do LSTM, CNN, and LLM models incorporating contextual information perform in binary automatic content moderation tasks for German newspaper comments compared to previous classification approaches?}

To explore this question, we use the One Million Posts Corpus \cite{schabus2017}, which comprises 1,000,000 posts from the Austrian newspaper Der Standard.\footnote{\url{https://www.derstandard.at/}, last accessed 2025-02-06}
Our findings show that CNN and LSTM models enhanced with contextual information achieve competitive performance compared to state-of-the-art transformer-based models such as BERT \cite{AssenmacherNMSR21,yadav2021,pachinger2024,keller2025}. However, we also demonstrate that incorporating contextual information does not significantly improve the zero-shot classification performance of ChatGPT 3.5 Turbo in terms of accuracy and F1-score. 

The remainder of this paper is structured as follows: Section \ref{sec:relatedwork} provides a review of the relevant literature and theoretical foundations, defining key concepts and presenting a taxonomy of existing methods in the field. Section \ref{sec:methods} outlines the methodology, including data preprocessing, the architectures of the CNN and LSTM models, and the prompt configurations used for ChatGPT. Section \ref{sec:results} presents the results, comparing our findings to existing studies. Finally, Section \ref{sec:conclusion} concludes the paper by summarizing key insights and highlighting directions for future research.

\section{Background \& Related Work\label{sec:relatedwork}}

\subsection{Hate Speech, Offensive Language, and Platform Guidelines}

In line with existing literature, we define \textit{automatic content moderation} as the use of computational decision programs to determine whether user-generated content (e.g., a post) complies with platform-specific rules. Based on this assessment, an appropriate moderation action is taken (e.g., the post is removed) \cite{ribeiro2023}. In contrast, manual content moderation refers to human moderators making these decisions instead of an automated system.

In the literature, several terms are closely linked to content moderation decisions, including \textit{hate speech} \cite[p. 85]{fortuna2018}, abusive language, cyberbullying, and \textit{offensive language} \cite{waseem2017}. Understanding these concepts is essential for analyzing recent research and, in particular, the annotation guidelines used for manual labeling in content moderation studies \cite{fortuna2018,waseem2017}.

Fortuna \& Nunes \cite{fortuna2018} provide an overview of various hate speech definitions, highlighting key aspects: (1) targeting specific groups based on characteristics like ethnicity or religion, (2) intent to incite violence or hatred, (3) language that demeans or attacks these groups, and (4) challenges in distinguishing hate speech from humor, particularly on platforms like Facebook.
Waseem et al. \cite{waseem2017} define abusive language along two dimensions: (1) whether it targets individuals or generalized groups, and (2) whether the abuse is overt or implicit. Explicit abuse includes clear slurs, while implicit abuse is often obscured by sarcasm or ambiguous language, making detection more challenging. 
Since hate speech is a subset of abusive language, broader definitions such as offensive language may be more applicable in content moderation. Zampieri et al. \cite{zampieri2019} define offensive language as ``any form of non-acceptable language (profanity) or a targeted offense, whether veiled or direct.'' 

Hate speech and offensive language are just one part of content moderation, which in newspaper forums often follows platform-specific rules. For example, Der Standard moderates not only for offensiveness but also for relevance, spam, personal data, and legal compliance. \footnote{Der Standard Community Guidelines, \url{https://www.derstandard.at/communityrichtlinien}, last accessed 2025-02-06} This work explores how machine learning can be trained to reflect such platform-specific guidelines for more effective automated moderation.

\subsection{Datasets for Automatic German Newspaper Moderation}

The One Million Posts Corpus \cite{schabus2017} is a German-language dataset featuring user comments from the Austrian newspaper DerStandard, annotated for content moderation. It includes 1,000,000 binary-labeled posts (online/offline) and 11,773 multi-labeled posts, categorized by professional moderators based on factors such as sentiment, relevance (on-/off-topic), inappropriate language, discrimination, feedback, personal stories, and argumentation. The dataset provides rich contextual information, including comment structure, timestamps, article metadata, and user interactions. 
The dataset is widely used in research on automated moderation in German newspaper forums \cite{schabus2017,AssenmacherNMSR21,yadav2021}. Given its large-scale binary-labeled data, it serves as a robust benchmark for training and evaluating models in this study.

The Rheinische Post dataset \cite{AssenmacherNMSR21} consists of user comments from the Rheinische Post website,\footnote{\url{https://rp-online.de/}, last accessed 2025-02-06} annotated by moderators and crowdworkers using a custom labeling schema that includes offensive categories (e.g., sexism, racism, insults) and organizational labels (e.g., advertisement, meta-discussions). Unlike the One Million Posts Corpus, the Rheinische Post dataset lacks contextual information for the postings. The NDR Dataset \cite{yadav2021} includes labeled comments from NDR news articles,\footnote{Dataset currently not accessible, \url{https://www.ndr.de/}, last accessed 2025-02-06} The dataset has only binary labels (offline/online) for each comment and no contextual details, besides the article title and article URL. 
AustroTox contains Austrian news forum comments annotated for offensiveness, including token-level spans marking vulgar language and targets of offensive statements. The dataset contains the title of the articles as additional contextual information \cite{pachinger2024}. HOCON34k comprises over 34,000 German newspaper comments labeled for hate speech and the absence or presence of enough context and contains no additional contextual information \cite{keller2025}. GERMS-AT includes 8,000 user comments from Austrian Newspaper forums, annotated on a 0–4 scale for sexist and misogynistic content, and has no contextual information beyond that \cite{krenn2024}. 

\subsection{Taxonomy of Content Moderation Methods}

Hate speech and offensive language detection are typically framed as supervised machine learning tasks, requiring large manually labeled datasets \cite{AssenmacherNMSR21}. Based on literature reviews on content moderation \cite{alkomah2022,malik2024,mullah2021} the approaches can be categorized into three primary groups: 
\paragraph{(1) Traditional (Shallow) Methods.}
Shallow models rely on conventional word representation techniques like TF-IDF and n-grams \cite{malik2024}. Sentiment lexicons and linguistic features can enhance classification accuracy. Popular classifiers include Support Vector Machines (SVM), Naïve Bayes (NB), logistic regression, decision trees, and K-Nearest Neighbors \cite{mullah2021}.
\paragraph{(2) Advanced Methods.}
Word embeddings generate vectorized representations that capture semantic relationships between words \cite{malik2024}. Common methods include word2vec \cite{mikolav2013}, fasttext \cite{joulin2017}, and GloVe \cite{pennington2014}. These embeddings are used in combination with traditional classifiers or deep neural networks, improving performance compared to shallow methods.

Deep learning models process input using either traditional feature encoding (e.g., TF-IDF) or pre-trained embeddings. Established architectures include:
\begin{itemize}
    \item Long Short-Term Memory (LSTM) models: Introduced by Hochreiter \& Schmidhuber \cite{hochreiter1997}, LSTMs effectively retain long-term dependencies. They have been applied in German newspaper comment moderation \cite{schabus2017}.
    \item Convolutional Neural Networks (CNNs): CNNs have been successfully applied to hate speech detection in English datasets \cite{alkomah2022} but remain underexplored in German-language moderation.
    \item Transformer-based Models: Introduced in Attention Is All You Need \cite{vaswani2017}, finetuned transformers, based on e.g.,  BERT, ELECTRA, or ALBERT, outperform earlier deep learning models in hate speech classification \cite{malik2024,Saumya2024,Jose2025}.
\end{itemize}
\paragraph{(3) Generative Pretrained Transformer-based LLMs.}
Generative Large Language Models (LLMs), such as ChatGPT 3.5, leverage transformers and allow zero-shot or few-shot learning via prompt engineering \cite{dehghan2024,li2024,Lu2025,Mnassri2024,Jaremko2025,chiu2022,pachinger2024}. OpenAI's API enables automated moderation pipelines; on the other hand, open source alternatives like Mistral 7B provide local deployment options.\footnote{\url{https://mistral.ai/en/news/announcing-mistral-7b}, last accessed 2025-02-07} Effective prompt design is crucial for achieving optimal classification performance \cite{li2024,pachinger2024}.

\subsection{Offensive language detection in German newspaper comments}

Despite the growing importance of automatic moderation, research on German-language newspaper comment moderation remains limited, particularly when focusing on a binary online/offline decision \cite{yadav2021}. Instead, many studies center on detecting specific categories such as offensive or abusive language, rather than addressing content moderation as a whole \cite{AssenmacherNMSR21,schabus2017}. This distinction is crucial, as different definitions and conceptual scopes lead to variations in annotation guidelines and dataset labeling practices in this research field \cite{fortuna2018}.   

\newcolumntype{k}{>{\hsize=.33\hsize}X}

\begin{table}[!ht]
\centering
\scriptsize
\resizebox{\textwidth}{!}{%
\begin{tabularx}{\textwidth}{kkk}
\toprule
\textbf{Paper}                           & \textbf{Classifier}     & \textbf{Word Representation}   \\ \midrule
\multirow{6}{*}{Schabus et al. 2017 \cite{schabus2017}}     & Support Vector Machine  & Countvectorizer (Bag of Words) \\
                                         & Multinomial Naive Bayes & Countvectorizer (Bag of Words) \\
                                         & Support Vector Machine  & naive Bayes log-count ratios   \\
                                         & Bag of Cluster          & word2vec                       \\
                                         & Support Vector Machine  & doc2vec                        \\
                                         & LSTM                    & pre-trained embedding          \\ \midrule
\multirow{7}{*}{Assenmacher et al. 2021 \cite{AssenmacherNMSR21}} & Multinomial N. Bayes    & tf-idf / fasttext              \\
                                         & Logistic Regression           & tf-idf / fasttext              \\
                                         & Gradient Boosted Trees  & tf-idf / fasttext              \\
                                         & AutoML                  & tf-idf / fasttext              \\
                                         & BERT     & /                              \\ \midrule
\multirow{6}{*}{Yadav \& Milde 2021 \cite{yadav2021}}     & Logistic Regression     & Countvectorizer (Bag of Words) \\
                                         & Logistic Regression     & doc2vec                        \\
                                         & Neural Network          & doc2vec                        \\
                                         & BERT                    & /                              \\ \midrule
\multirow{5}{*}{Pachinger et al. 2024 \cite{pachinger2024}} & BERT                    & /                              \\
                                         & Electra                 & /                              \\
                                         & GPT-3.5                 & /                              \\
                                         & GPT-4                   & /                              \\
                                         & LeoLM          & /                              \\
                                         & Mistral          & /                              \\ \midrule
                                         
Krenn et al. 2024 \cite{krenn2024}       & BERT                    & /                              \\ \midrule
Keller et al. 2025 \cite{keller2025}     & BERT                    & /                              \\
\bottomrule
\end{tabularx}
}

\caption{Overview of classifiers and word representations used in studies on German newspaper comment moderation.}
\label{tab:lit}
\end{table}

There are six major papers in the scientific literature \cite{schabus2017,AssenmacherNMSR21,yadav2021,pachinger2024,krenn2024,keller2025}; Table \ref{tab:lit} gives an overview of the classifiers and word representations used in the papers. Notably, Yadav \& Milde \cite{yadav2021} incorporate contextual information by using both the article title and comment, as well as splitting data by topic. Pachinger et al. \cite{pachinger2024} use the article title as contextual information in their experiments, while Keller et al. \cite{keller2025} consider the absence of enough contextual information during their data annotation process.

\section{Methods\label{sec:methods}}

\subsection{Data Overview and Preprocessing}

We first give an overview of the data recorded in the One Million Posts Corpus in Table \ref{tab:omp_posts} for posts and Table \ref{tab:omp_articles} for articles. We use the columns \texttt{Path}, \texttt{Title} of the article, \texttt{Headline} and \texttt{Body} of the post (merged into a \texttt{Comment} variable) as well as two engineered features $R_o^s$ and $R_o^f$ described further down below to predict the \texttt{Status} of the post, i.e. whether it is still ``online'' or was ``deleted'' by a moderator.

\newcolumntype{d}{>{\hsize=.7\hsize}X}
\newcolumntype{e}{>{\hsize=.2\hsize}X}

\begin{table}[ht]
    \scriptsize
    \centering
    \def\arraystretch{1.2}
    \begin{tabularx}{\textwidth}{ldle}
        \toprule
         \textbf{Name} & \textbf{Description} & \textbf{Type} & \textbf{Example} \\
         \midrule
         \texttt{Post ID} & Identifier of the Post & \texttt{integer} & 81085\\
         \texttt{Article ID} & The ID of the article the comment was posted under & \texttt{integer} & 1212\\
         \texttt{Parent Post} & NULL for top-level comments, otherwise the ID of the parent comment & \texttt{integer or NULL} & 80997\\
         \texttt{User ID} & The (anonymized) user ID & \texttt{integer} & 7721\\
         \texttt{Headline} & The headline of the post & \texttt{text or NULL} & NULL\\
         \texttt{Body} & The main text of the post & \texttt{text or NULL} & Drei Packerl Karten á 36 Blatt pro Jahr\\
         \texttt{Time Stamp} & When the post was created & \texttt{date} & 2015-07-02 12:25:53.553\\
         \texttt{Positive Votes} & upvotes by other community members & \texttt{integer} & 0\\
         \texttt{Negative Votes} & downvotes by other community members & \texttt{integer} & 0\\
         \texttt{Status} & Whether the post is ``online'' or ``deleted'' & \texttt{text} & online\\
         \bottomrule
    \end{tabularx}
    \caption{The data recorded in the One Million Posts Corpus for each post.}
    \label{tab:omp_posts}
\end{table}

\newcolumntype{f}{>{\hsize=.4\hsize}X}
\newcolumntype{g}{>{\hsize=.6\hsize}X}
\begin{table}[ht]
    \scriptsize
    \centering
    \def\arraystretch{1.2}
    \begin{tabularx}{\textwidth}{lflg}
        \toprule
         \textbf{Name} & \textbf{Description} & \textbf{Type} & \textbf{Example} \\
         \midrule
         \texttt{Article ID} & Identifier of the article & \texttt{integer} & 1212\\
         \texttt{Path} & Breadcrumbs to the article & \texttt{text} & Newsroom/Panorama/Chronik\\
         \texttt{Date} & Date of Publishing & \texttt{date} & 2015-07-02 05:30:00.00\\
         \texttt{Title} & Title of the article & \texttt{text} & Damenstift in Innsbruck: Adelsfrauen, die täglich für den Kaiser beten\\
         \texttt{Body} & Content of the article & \texttt{text} & \textit{omitted for brevity}\\
         \bottomrule
         \vspace*{2px}
    \end{tabularx}
    \caption{The data recorded in the One Million Posts Corpus for each article.}
    \label{tab:omp_articles}
\end{table}

There is a large class imbalance between ``online'' and ``deleted'' posts in the One Million Posts Corpus, with 1,011,773 labeled as ``online'' and 62,320 labeled as ``deleted''. This can lead to a worse classification performance of the resulting models, so following the recommendations of \cite{raschka2022}, we balance the data by randomly dropping ``online'' comments, resulting in 124,640 total entries, 62,320 ``online'' and 62,320 ``deleted''.


The \texttt{Path}, \texttt{Title} and \texttt{Comment} variables are text variables, so we will need further preprocessing on them. Following the suggestion of \cite{kang2020}, we make use of encoding, cleaning, tokenization, stop word removal, and lemmatization. After importing, the data is further cleaned by converting all text to lowercase and removing non-alphabetic characters. After that, lemmatization is done. Lemmatization brings the words to a more general form. Then the text is tokenized into single words. In case a model receives multiple text variables as input, the specific variables are concatenated with special tokens. ``LINK'' for the topic path,``TITEL'' for the title of the article, and ``KOMMENTAR'' for the user comment. 

We also construct two additional features from the \texttt{User ID}, a simple online ratio $R_o^s$, and the full online ratio $R_o^f$, as follows:
\begin{align*}
    R_o^s&\coloneqq \frac{C_\text{online}^\text{train}}{C^\text{train}},\\
    R_o^f&\coloneqq \frac{C_\text{online}^\text{train}+C_\text{online}^\text{ds}}{C^\text{train}+C^\text{ds}},
\end{align*}
where $C^\text{train}$ is the number of comments of the user in the training set, $C^\text{ds}$ is the number of comments of the user in the part of the dataset lost to downsampling, $C_\text{online}^\text{train}$ is the number of online comments of the user in the training set, $C_\text{online}^\text{ds}$ is the number of online comments of the user in the part of the dataset lost to downsampling. $R_o^s$ therefore represents the percentage of user comments in the downsampled dataset that are still online, while $R_o^f$ represents the same percentage in the whole dataset. Note that $C^\text{ds}$ and $C_\text{online}^\text{ds}$ do not use any data points from the test or validation sets to prevent any form of data leakage.
\newpage
\subsection{Architectures}

\begin{figure}[ht!]
    \centering
    \includegraphics[width=0.95\linewidth]{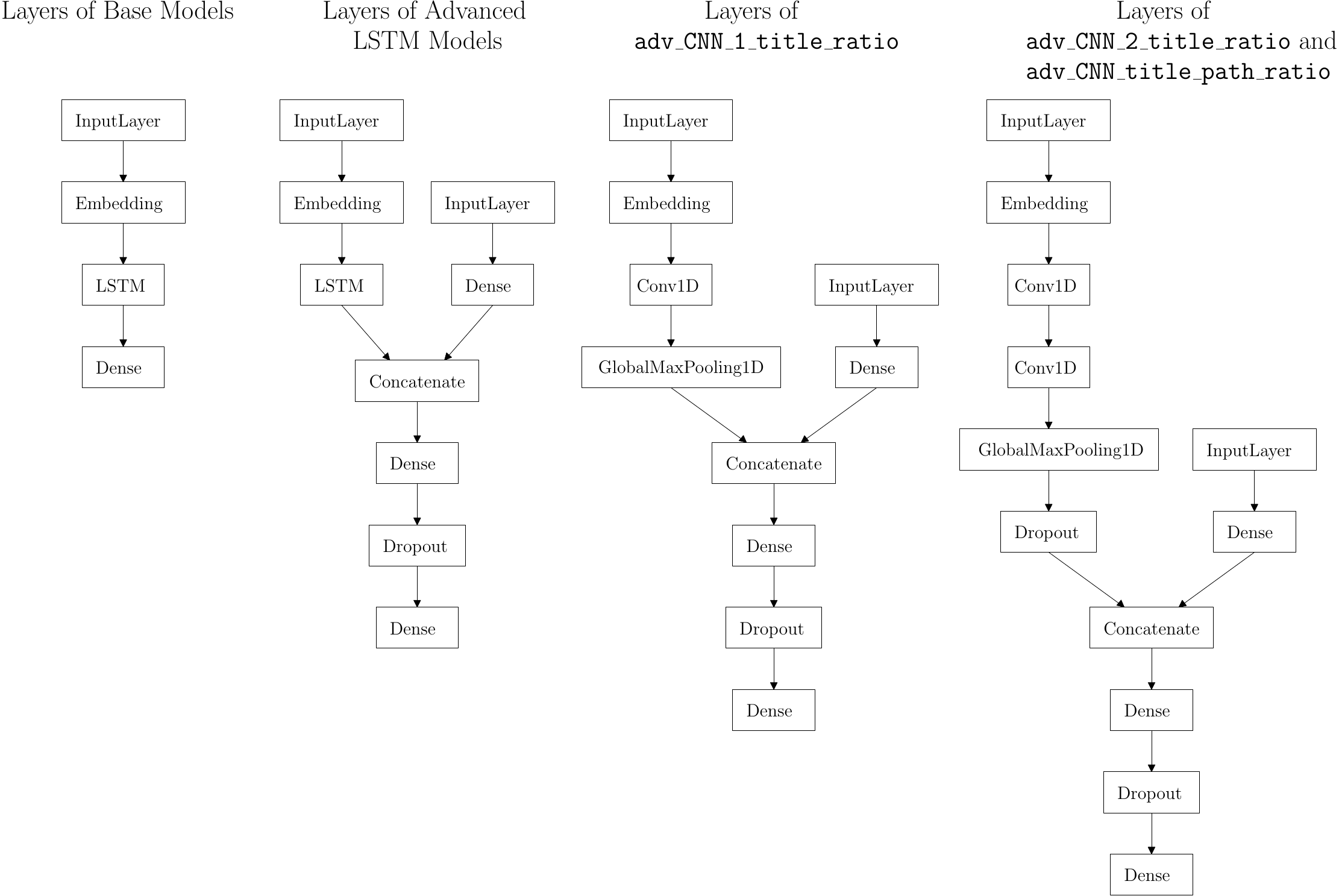}
    \vspace*{-20px}
    \caption{The different deep learning architectures.}
    \label{fig:arch}
\end{figure}

The model architectures used in this paper can be categorized into three groups: Traditional shallow methods as baselines, deep learning based approaches and generative pretrained transformers. A graphical comparison of the deep learning architectures is shown in Figure \ref{fig:arch}.

\paragraph{Traditional Shallow Methods.}
Traditional shallow models are utilized as baseline models, with no additional context information, thus predicting the binary classification task solely with the cleaned text of the user comment.  

After the data preprocessing explained in the last section, it is crucial to transform the text into a numerical representation. For this a Countvectorizer is used and on top of that a traditional classifier. We use a multinomial naive Bayes and logistic regression.

\paragraph{Deep Learning Methods.}
The deep learning methods additionally use LSTM and CNN architectures. As embeddings we use pre-trained word embeddings from fasttext \cite{joulin2017}. We use three models with simple LSTM architectures, three with more advanced LSTM architectures, and three with CNN architectures, which all differ in the inputs they take to make their predictions. These are summarized in Table \ref{tab:model_inputs}.

\begin{table}[ht]
    \scriptsize
    \centering
    \def\arraystretch{1.2}
    \begin{tabularx}{\textwidth}{ff}
        \toprule
         \textbf{Name} & \textbf{Inputs Used} \\
         \midrule
         \texttt{base\_LSTM} & \texttt{Comment}\\
         \texttt{base\_LSTM\_title} & \texttt{Comment}+\texttt{Title}\\
         \texttt{base\_LSTM\_title\_path} & \texttt{Comment}+\texttt{Title}+\texttt{Path}\\
         \texttt{adv\_LSTM\_Title\_simple\_ratio} & \texttt{Comment}+\texttt{Title}+$R_o^s$\\
         \texttt{adv\_LSTM\_Title\_ratio} & \texttt{Comment}+\texttt{Title}+$R_o^f$\\
         \texttt{adv\_LSTM\_Title\_Path\_ratio} & \texttt{Comment}+\texttt{Title}+\texttt{Path}+$R_o^f$\\
         \texttt{adv\_CNN\_1\_title\_ratio} & \texttt{Comment}+\texttt{Title}+$R_o^f$\\
         \texttt{adv\_CNN\_2\_title\_ratio} & \texttt{Comment}+\texttt{Title}+$R_o^f$\\
         \texttt{adv\_CNN\_title\_path\_ratio} & \texttt{Comment}+\texttt{Title}+\texttt{Path}+$R_o^f$\\
         \bottomrule
         \vspace*{2px}
    \end{tabularx}
    \vspace*{-20px}
    \caption{The inputs for the different deep learning models.}
    \label{tab:model_inputs}
\end{table}
\paragraph{Generative Pretrained Transformer-based LLMs.}
For the generative transformer-based models we use the OpenAI API \cite{openaiapi} and the model ``ChatGPT: GPT-3.5-Turbo'' to do the experiments, following the discussion in Li et al. \cite{li2024}. Building on recent findings in prompt design, we focus on providing richer contextual information to improve classification. Comments are input as raw text. Hyperparameters are tuned following Li et al., with temperature set to 0 for determinism and top-p to 0.95. To assess the impact of context, we test multiple prompt variations, starting from the following base prompt:\footnote{English translation: \texttt{\scriptsize You are a forum moderator and are responsible for moderating comments under a newspaper article. Make a prediction for the moderation decision whether the comment should remain online ``0'' or be taken offline ``1''. The comment is: ``[}\textit{\scriptsize comment}\texttt{\scriptsize]''. Reply exclusively in Json format \{``Moderation decision'': prediction\}}}
\begin{quote}\small
    \texttt{Du bist ein Forenmoderator und dafür zuständig, Kommentare unter einem Zeitungsartikel zu moderieren. Mache eine Prediction zur Moderations\-entscheidung ob das Kommentar Online bleiben soll ``0'' oder Offline ge\-nommen werden soll ``1''. Das Kommentar ist: ``[}\textit{comment}\texttt{]''. Antworte aus\-schließlich im Json Format \{``Moderationsentscheidung'': prediction\}}
\end{quote}
In total, we use 7 different variants, summarized in Table \ref{tab:transformer_variants}, including the use of the newspaper's forum rules.\footnote{\url{https://about.derstandard.at/agb/\#Forum}, last accessed 2025-02-11}

\newcolumntype{h}{>{\hsize=.6\hsize}X}
\begin{table}[ht]
    \scriptsize
    \centering
    \def\arraystretch{1.2}
    \begin{tabularx}{\textwidth}{gh}
        \toprule
         \textbf{Name} & \textbf{Information in Prompt} \\
         \midrule
         \texttt{GPT\_base} & base prompt only\\
         \texttt{GPT\_mod\_title} & base prompt modified to include the article title\\
         \texttt{GPT\_mod\_title\_strength} & base prompt + article title + request for output of strength of prediction\\
         \texttt{GPT\_mod\_title\_ratio} & base prompt + article title + $R_o^f$\\
         \texttt{GPT\_mod\_title\_path} & base prompt + article title + path\\
         \texttt{GPT\_mod\_title\_erklaerung} & base prompt + article title + request for explanation of decision\\
         \texttt{GPT\_mod\_title\_forenregeln\_kurz\_erklaerung} & base prompt + article title + short summary of forum rules + request for explanation of decision\\
         \bottomrule
         \vspace*{2px}
    \end{tabularx}
    \caption{The different prompt configurations of the GPT models.}
    \label{tab:transformer_variants}
\end{table}

\section{Results\label{sec:results}}

This section presents the evaluation of various machine learning models for binary classification tasks, using performance metrics such as accuracy, F1-score, precision, recall, and AUROC, as outlined by \cite{raschka2022}. The models assessed include advanced LSTM and CNN architectures, simpler classifiers such as naive Bayes and logistic regression, and models based on GPT prompts.

\begin{table}[htbp]
\centering
\scriptsize
\begin{tabularx}{\textwidth}{g c c c c c c}
\toprule
\textbf{Model} & \textbf{Accuracy} & \textbf{AUROC} & \textbf{F1-Score} & \textbf{Precision} & \textbf{Recall} & \textbf{Missing Answer} \\ \midrule
\texttt{adv\_LSTM\_Title\_path\_ratio} & 0.733 & 0.809 & 0.713 & 0.734 & 0.692 & / \\ 
\texttt{adv\_LSTM\_Title\_ratio} & 0.729 & 0.808 & 0.718 & 0.714 & 0.723 & / \\ 
\texttt{adv\_CNN\_1\_title\_ratio} & 0.728 & 0.804 & 0.699 & 0.741 & 0.663 & / \\ 
\texttt{adv\_CNN\_title\_path\_ratio} & 0.726 & 0.804 & 0.703 & 0.730 & 0.681 & / \\ 
\texttt{adv\_CNN\_2\_title\_ratio} & 0.726 & 0.802 & 0.701 & 0.733 & 0.673 & / \\ 
\texttt{adv\_LSTM\_Title\_simple\_ratio} & 0.725 & 0.802 & 0.725 & 0.695 & 0.757 & / \\ 
\texttt{base\_LSTM\_title\_path} & 0.703 & 0.777 & 0.698 & 0.678 & 0.720 & / \\ 
\texttt{base\_LSTM\_title} & 0.699 & 0.772 & 0.687 & 0.683 & 0.693 & / \\ 
\texttt{naive\_bayes} & 0.678 & 0.743 & 0.679 & 0.677 & 0.681 & / \\ 
\texttt{logistic\_regression} & 0.663 & 0.726 & 0.656 & 0.670 & 0.643 & / \\ 
\texttt{base\_LSTM} & 0.661 & 0.728 & 0.667 & 0.629 & 0.710 & / \\ 
\texttt{GPT\_mod\_title\_path} & 0.635 & / & 0.651 & 0.609 & 0.700 & 75 \\ 
\texttt{GPT\_base} & 0.632 & / & 0.652 & 0.601 & 0.712 & 56 \\ 
\texttt{GPT\_mod\_title\_ratio} & 0.631 & / & 0.643 & 0.605 & 0.687 & 158 \\ 
\texttt{GPT\_mod\_title\_erklaerung} & 0.629 & / & 0.654 & 0.597 & 0.725 & 44 \\ 
\texttt{GPT\_mod\_title} & 0.626 & / & 0.646 & 0.598 & 0.702 & 63 \\ 
\texttt{GPT\_mod\_title\_forenregeln\_ kurz\_erklaerung} & 0.613 & / & 0.672 & 0.571 & 0.816 & 53 \\ 
\texttt{GPT\_mod\_title\_strength} & 0.606 & / & 0.673 & 0.563 & 0.835 & 39 \\ 
\bottomrule
\vspace*{2px}
\end{tabularx}
\vspace*{-10px}
\caption{Model Performance Metrics}
\label{tab:model_performance_metrics}
\end{table}

\paragraph{Model Performance Overview.}
The results summarized in Table \ref{tab:model_performance_metrics} indicate that the advanced LSTM and CNN models consistently outperform simpler approaches. The \texttt{adv\_LSTM\_Title\_path\_ratio} model achieves the highest accuracy (0.733) and AUROC (0.809). Conversely, simpler models such as logistic regression and naive Bayes perform at a lower level, with accuracy scores of 0.663 and 0.678, respectively. The GPT-based models exhibit the poorest overall performance, with accuracy scores ranging from 0.606 to 0.635, and high rates of ``missing answer'' -- instances where the model fails to classify comments due to incorrect formats or refusal to provide a response.

\paragraph{Accuracy and the Role of Contextual Information.}
As seen in Table \ref{tab:model_performance_metrics}, adding contextual variables significantly enhances the performance of self-trained models, whereas this does not hold for GPT-based models. For example, \texttt{base\_LSTM}, which uses only cleaned comment text as input, is outperformed by \texttt{base\_LSTM\_title} and \texttt{base\_LSTM\_title\_path}, which incorporate the article title and topic path as additional input features. Interestingly, GPT models do not show similar improvements when contextual information is added, indicating that the increased input complexity incurs additional computational costs without a corresponding increase in accuracy.


\begin{figure}[ht]
    \centering
    \includegraphics[width=0.6\linewidth]{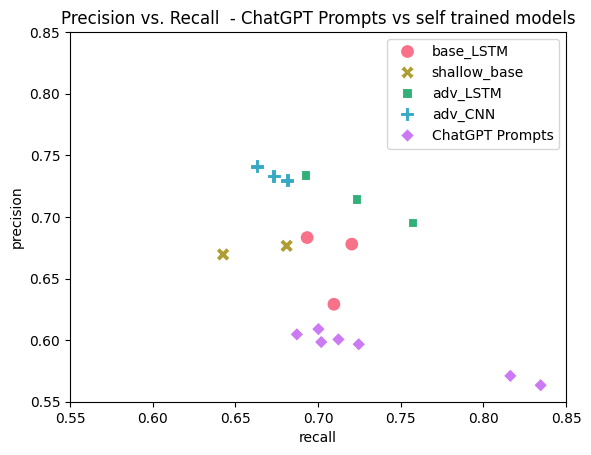}
    \vspace*{-10px}
    \caption{Precision-Recall comparison of the different models.}
    \label{fig:precision_recall}
\end{figure}

\paragraph{AUROC and F1-Score Analysis.}
The AUROC values for the models mirror the trends observed in accuracy. The advanced LSTM models demonstrate slightly superior AUROC scores compared to other models. In terms of F1-score, the advanced LSTM model using the simple ratio input achieves the highest score. Among the GPT models, \texttt{GPT\_title\_strength} achieves a relatively high recall but at the expense of precision, leading to a higher rate of false positives. These findings suggest that such models may be more suitable for human-assisted moderation systems, where maximizing recall is preferred over precision.

\paragraph{Precision and Recall Comparison.}
Figure \ref{fig:precision_recall} presents a comparison of precision and recall, revealing that the GPT-based models exhibit high recall but lower precision, implying a higher false positive rate. In contrast, the self-trained models demonstrate a more balanced performance between precision and recall, with the advanced LSTM and CNN models outperforming other approaches. Models incorporating user history, in the form of the Online Ratio variable, show the best performance, while the inclusion of the \texttt{Path} variable contributes minimally to performance improvements.

\paragraph{Comparison of Self-Trained and GPT Models.}
Overall, the self-trained models outperform GPT-based models, particularly in terms of accuracy and F1-score. The advanced LSTM and CNN models exhibit superior results, with the \texttt{adv\_LSTM\_Title\_path\_ratio} model achieving the highest accuracy and AUROC values. Although computationally less expensive, simpler classifiers such as logistic regression and naive Bayes outperform the \texttt{base\_LSTM} model but are consistently outperformed by more advanced models such as \texttt{base\_LSTM\_title} and \texttt{base\_LSTM\_title\_path}. It is also worth noting that there are no significant differences in performance between the advanced LSTM and CNN architectures.

\paragraph{Comparison with Previous Literature.}
When compared to previous work in the field, such as \cite{AssenmacherNMSR21}, \cite{pachinger2024},\cite{keller2025}  and \cite{yadav2021}, the self-trained models presented in this study exhibit competitive performance. While Assenmacher’s best model (BERT) achieves a superior AUROC score (0.914), the highest-performing model in this study (\texttt{adv\_LSTM\_Title\_path\_ratio}) achieves a strong AUROC of 0.809. Notably, when comparing models based on professional moderator labels, the advanced CNN and LSTM models developed in this study outperform the professional-moderator-based models from \cite{AssenmacherNMSR21}. Similarly, compared to \cite{yadav2021}, our best model achieves higher F1-score, precision, and recall, although it underperforms in terms of accuracy.

\section{Conclusion\label{sec:conclusion}}
This study explored the potential of incorporating contextual information into automatic content moderation systems for German-language newspaper comments. By integrating context variables such as user posting history, forum rules, and article titles into LSTM, CNN, and large language models like ChatGPT 3.5 Turbo, the results show that LSTM and CNN models with context can compete with state-of-the-art transformer models like BERT in terms of performance.

Another finding is that adding context to GPT-3.5 Turbo prompts did not significantly improve classification accuracy, with simpler, context-free prompts performing comparably, underscoring the need to balance model complexity and cost. Additionally, the study highlights the potential of context-enriched models to improve moderation.
A limitation of our work is the continuous advancement of LLMs, meaning that more recent models may improve the performance and integration of contextual information.

Future research could explore adding more detailed contextual data, fine-tuning transformer models, and developing models capable of explaining their classification decisions, which could be important for legal and ethical considerations. The cost-effectiveness of using contextual models and maintaining fairness in variables like the Online Ratio are critical concerns for further exploration.

\bibliographystyle{splncs04}
\bibliography{references}

\end{document}